\documentclass[lettersize,journal]{IEEEtran}
\usepackage{amsmath,amsfonts}
\usepackage{algorithmic}
\usepackage{algorithm}
\usepackage{array}
\usepackage{textcomp}
\usepackage{stfloats}
\usepackage{url}
\usepackage{verbatim}
\usepackage{graphicx}
\usepackage{cite}
\usepackage{booktabs}
\usepackage{multirow}
\usepackage{tabularx}
\usepackage{subcaption}
\usepackage{caption}
\usepackage{float}
\usepackage[section]{placeins}
\begin{document}

\title{E2E Parking Dataset: An Open Benchmark for End-to-End Autonomous Parking}


\author{Kejia Gao, Liguo Zhou*, Mingjun Liu, Alois Knoll,~\IEEEmembership{Fellow,~IEEE,}
\thanks{Kejia Gao, Liguo Zhou and Alois Knoll are with the Chair of Robotics, Artificial Intelligence and Real-time Systems, Technical University of Munich, Garching bei München, Germany.}
\thanks{Liguo Zhou is with the School of Computer Science, Nanjing University of Information Science and Technology, Nanjing, China.}
\thanks{Mingjun Liu is with the Anhui Engineering Research Center for Intelligent Computing and Information Innovation, Fuyang Normal University, Fuyang, China.}
\thanks{* Liguo Zhou is the corresponding author (liguo.zhou@tum.de).}}

\markboth{Manuscript}%
{Shell \MakeLowercase{\textit{et al.}}: A Sample Article Using IEEEtran.cls for IEEE Journals}


\maketitle

\begin{abstract}

While traditional autonomous driving methods with multi-stage pipelines suffer from lengthy processes, error accumulations and maintenance difficulties, the end-to-end method is designed to map the data of multiple sensors directly into motion control commands, with high flexibility, efficiency and generalization. Therefore, the end-to-end model has shown great potential in autonomous driving. Due to the low speed, low risk, and low complexity characteristics of autonomous parking scenarios, end-to-end methods can be applied to autonomous parking systems earlier. However, the lack of publicly available datasets limits reproducibility and benchmarking in the early research stage of end-to-end autonomous parking. While prior work introduced a visual-based parking model and a pipeline for data generation, training and closed-loop test, the dataset itself was not released. To bridge this gap, we work on creating large end-to-end autonomous parking datasets in CARLA based on the prior work ‘E2E Parking’. Keyboard control is replaced by Handle Controller to improve usability, efficiency, and operational precision. During the iterative process of dataset generation, we evaluate the effect of different factors on the parking performance of the controlled vehicle, including diverse scenes generated by multiple random seeds, the position of the roadside object's shadow dependent on weather setting, dataset size, initial learning rate and training epochs. We recommend generating at least 2 scenes for each parking slot with different random seeds, where 8 trajectories with different initial positions are collected for each scene. Weather settings should be modified to make the dataset include scenes with shadow projected on the target slot. Experiments demonstrate that an initial learning rate of $7.5 \times 10^{-5}$ and epochs set as around 150 can make the model perform better. After several iterations, we are able to open-source a high-quality dataset for end-to-end autonomous parking. Using the original transformer-based model, we achieve an overall success rate of 85.16\% with lower average position and orientation errors (0.24 meters and 0.34 degrees). The dataset can be found in \url{https://github.com/KejiaGao/e2e-parking-carla-dataset}. 
\end{abstract}


\section{Introduction}
\IEEEPARstart{T}{he} field of autonomous driving has advanced significantly, with a focus on improving autonomous parking. Traditional multi-stage pipelines, including environment perception, mapping, path planning, and motion control, often suffer from sensor noise, model uncertainties, and rigid handcrafted rules, leading to error accumulation and poor adaptability in complex environments.

End-to-end method has emerged as a potential alternative by directly mapping sensor data to motion control commands, streamlining the pipeline, reducing reliance on handcrafted features, and improving adaptability. This approach has shown success in lane keeping, object detection, and collision avoidance. Inspired by this, previous work proposed an end-to-end parking method integrating visual inputs and motion states to generate control commands.

However, a key challenge in this domain is the lack of publicly available datasets for training and evaluation. While Yang et al.~\cite{yang2024e2e} introduced a model and shared a pipeline ranging from data generation to training and closed-loop test, their dataset remains unreleased and the pre-trained model they have published has only an overall success rate of around 75\%. To address this gap, we create and open-source an end-to-end autonomous parking dataset based on their environmental setup.

In this paper, we present the iterative process of creating a parking dataset that captures a wide variety of parking scenarios. Using the model architecture proposed by the original authors, we trained the model on the dataset and achieved an overall success rate of 85.16\%, which can replicate and exceed the performance of the released pre-trained model. Furthermore, it is comparable to the original authors’ result of 91.41\% from their unpublished model. Additionally, we observed significant reductions in both average position and orientation errors (0.24 meters and 0.34 degrees) compared to the baseline (0.30 meters and 0.87 degrees), indicating improved parking precision.

Our work not only provides a valuable dataset for the research community but also demonstrates the effectiveness of the original model on a brand new and diverse dataset. To encourage further research and development in the field of end-to-end autonomous parking including optimizing algorithm and iterating dataset, we release the latest dataset and comparison experiment results during the dataset iteration process for public access. The main contributions of this paper include:
\begin{itemize}
    \item The creation and open-sourcing of a new, high-quality end-to-end autonomous parking dataset based on CARLA, addressing the lack of available data in this domain.
    \item The application of the original end-to-end autonomous parking model to this new dataset, achieving competitive results in terms of Target Success Rate, Average Position Error and Average Orientation Error.
    \item The release of iteration details for dataset creation, providing a foundation for future research in autonomous parking.
\end{itemize}

\section{Related Works}

In this part, the development of Transformer model and BEV model used in autonomous driving will be reviewed, and based on this, the application of these algorithms in the field of automatic parking will be introduced.

\subsection{Transformer Model}

Transformer models have revolutionized AI since their introduction in the paper ``Attention Is All You Need"~\cite{vaswani2017attention} by Vaswani et al. in 2017. It was first applied to Natural Language Processing (NLP) where it achieved significant success, such as BERT~\cite{devlin2019bert}, GPT~\cite{brown2020language} and T5~\cite{raffel2020exploring}. In the field of computer vision (CV), Vision Transformers (ViTs)~\cite{dosovitskiy2020image} are able to outperform CNNs in image classification tasks by capturing both global and local features and understanding image-wide context. DETR~\cite{carion2020end} is a successful end-to-end object detection framework with a Transformer encoder-decoder architecture.

Transformer models are also widely applied to autonomous driving. Transfuser~\cite{chitta2022transfuser} and InterFuser~\cite{shao2023safety} use transformers for multi-scale fusion of image and LiDAR features. BEVFusion~\cite{liu2023bevfusion} is a framework that combines camera and LiDAR features in the Bird's Eye View (BEV) space and uses transformers to fuse multi-modal data, promoting the understanding of the driving environment. UniAD~\cite{hu2023planning} utilizes four transformer decoder-based perception and prediction modules to establish end-to-end autonomy. ParkPredict+~\cite{shen2022parkpredict+} uses transformers for waypoint prediction in complicated parking environments. DriveTransformer~\cite{jia2025drivetransformer} is a unified transformer framework, introducing task parallelism, sparse representation, and streaming processing to enhance training stability and efficiency. TransParking~\cite{du2025transparking} is a vision-based transformer framework for end-to-end autonomous parking, using expert trajectory training to directly predict future coordinates. E2E Parking~\cite{yang2024e2e} sets up a pipeline for data generation, training and close-loop test based on CARLA simulator, implements an LSS-based transformer architecture for imitation learning, predicts control signals directly from RGB images and motion states, and achieves a target success rate of 91\%. As an extention, ParkingE2E~\cite{li2024parkinge2e} proposes a camera-based end-to-end parking model that fuses surround-view RGB images in BEV space by target query and predicts trajectory points using a transformer decoder, demonstrating reliability across various scenarios.

\subsection{BEV Representation}

The Bird’s-Eye View (BEV) is widely used in autonomous driving for tasks like semantic segmentation and object detection, providing a top-down perspective of the environment. Since 2D image features lack depth, LSS~\cite{philion2020lift} method learns depth distributions and uses cameras' intrinsic and extrinsic matrices to transform the frustum into BEV. Based on LSS, BEVDepth~\cite{li2023bevdepth} improves depth estimation with explicit supervision. Transformer models, as in Transfuser~\cite{chitta2022transfuser}, are used for BEV generation by fusing LiDAR and image features. ST-P3~\cite{hu2022st} is an end-to-end vision-based autonomous driving framework that leverages spatial-temporal BEV feature learning for perception, prediction and planning.

\subsection{Autonomous Driving Datasets}

The KITTI dataset~\cite{geiger2013vision} provides 6 hours of diverse real-world traffic scenarios with synchronized sensor data, including stereo cameras, a 3D laser scanner and GPS/IMU, supporting tasks like stereo vision, optical flow, and object detection. The nuScenes dataset~\cite{caesar2020nuscenes}, with a full autonomous vehicle sensor suite (6 cameras, 5 radars, 1 LiDAR), provides 1000 fully annotated 20s scenes with 3D bounding boxes for 23 classes, aiming at detection and tracking evaluation with novel metrics. The Waymo Open Dataset~\cite{sun2020scalability} consists of 1150 diverse, high-quality 20s scenes with synchronized and calibrated LiDAR and camera data and exhaustive 2D and 3D annotations, used for 2D and 3D detection and tracking tasks. NAVSIM~\cite{dauner2024navsim} is a benchmarking framework for non-reactive autonomous vehicle simulation, utilizing diverse datasets like nuPlan to evaluate end-to-end driving policies through standardized open-loop metrics, supporting tasks such as planning and trajectory prediction.

For autonomous parking, the PKLot Dataset~\cite{de2015pklot} contains 695,899 images from two parking lots with three camera views under various weather conditions, enabling benchmarking and evaluation of parking vacancy detection algorithms. SUPS~\cite{hou2022sups} is a simulated underground parking dataset featuring multi-sensor data (fisheye cameras, pinhole cameras, LiDAR, IMU, GNSS) with pixel-level semantic labels, applied to tasks such as SLAM, semantic segmentation, parking slot detection, 3D reconstruction and depth estimation. The Dragon Lake Parking (DLP) dataset~\cite{shen2022parkpredict+} provides high-resolution aerial video and detailed annotations of vehicles, cyclists, and pedestrians in a large parking lot, supporting research on trajectory analysis, object detection, tracking, and semantic segmentation.

Most of the related datasets are targeted at detection, classification, segmentation and planning, while the mainstream end-to-end autonomous parking methods primarily predict trajectory points. E2E Parking is one of the few studies that directly predict control signals. However, the original authors do not release their dataset, limiting further research on the algorithm, which serves as our motivation for developing a dataset for it.

\section{Method}

The innovative architecture~\cite{yang2024e2e} proposed by the original authors is continued in this work, where a neural network maps inputs to control command outputs.

\subsection{Problem Definition}

The system aims to achieve precise vehicle parking control using sensor measurements. The model is trained in a supervised learning framework with expert demonstration dataset $\mathcal{D} = \{\tau^n \mid n \in \{1,2,\dots,N\}\}$, where each of the N trajectories $\tau_n$ consists of sequential observations and control signals: $\tau^n = \left\{ \left( \mathcal{X}_t^n, \mathcal{C}_t^n \right) \right\}_{t=1}^{T^n}$. $\mathcal{X}_t^n$ includes RGB images of four cameras, target slot coordinates and ego vehicle's motion state in the $n$th trajectory at time step $t$, while $\mathcal{C}_t^n$ consists of values of throttle, brake, steering and gear state in the $n$th trajectory at time step $t$.

The learning objective is formulated as:
\begin{equation}
\arg\min_{\pi} \mathbb{E}_{(\mathcal{X},\mathcal{C})\sim\mathcal{D}} \left[ \mathcal{L} \left( \mathcal{C}, \pi(\mathcal{X}) \right) \right]
\end{equation}

where $\pi$ is the control policy mapping inputs to control signals, and $\mathcal{L}$ is the loss function.

\subsection{Inputs and Outputs}

\textbf{Inputs:} The model takes camera images, motion states (velocity and acceleration), and the target slot position. Four surrounding cameras (front, left, right, rear) are positioned to capture the environment. The target slot is initially set by the user and then continuously tracked via segmentation.

\textbf{Outputs:} The model predicts control signals for the next four steps with 0.1s interval. The outputs include normalized acceleration (\(\text{acc}_t \in [-1,1]\)), steering angle (\(\text{steer}_t \in [-1,1]\)), and gear selection (\(\text{gear}_t \in \{0,1\}\)). Inspired by Pix2Seq \cite{chen2021pix2seq}, control prediction is framed as a language modeling task, where signals are tokenized and predicted sequentially. Acceleration and steering are discretized into 201 values \([0,200]\), with 0 indicating full brake/left and 200 indicating full throttle/right. For gear state, 0 stands for forward gear and 200 means reverse gear. The sequence format is:

\begin{equation}
\begin{aligned}
S = [\text{BOS}, c_0, c_1, c_2, c_3, \text{EOS}] \\ 
c_n = [\text{acc}_n, \text{steer}_n, \text{gear}_n].
\end{aligned}
\end{equation}

\subsection{Network Architecture}

The architecture consists of BEV Generation, Feature Fusion and Control Prediction.

\textbf{BEV Generation}: Utilizing the LSS method\cite{philion2020lift}, multi-view images \( I \in \mathbb{R}^{3 \times H \times W} \) are processed by the backbone EfficientNet-B4 to extract features \( F_{\text{img}} \in \mathbb{R}^{C_i \times H_i \times W_i} \) to estimate depth distribution $D$. Feature frustums \( F_{\text{frus}} \in \mathbb{R}^{D \times C_i \times H_i \times W_i} \) are then generated, which are projected into a BEV voxel grid and merged to form the final BEV feature map \( F_{\text{bev}} \in \mathbb{R}^{C_b \times X_b \times Y_b} \).

\textbf{Feature Fusion}: The BEV feature, target slot, and ego motion are combined into a unified feature. By setting target slot grids as 1 and other grids as 0, the target feature \( F_{\text{target}} \in \mathbb{R}^{1 \times X_b \times Y_b} \) is projected into a BEV grid, concatenated with \( F_{\text{bev}} \), and processed through ResNet18 convolution layers to obtain \( F'_{\text{bev}} \in \mathbb{R}^{C'_b \times L} \), where $C'_b = C_b + 1$. Ego motion features \( F_{\text{ego}} \in \mathbb{R}^{3 \times L} \) are encoded by MLP, representing the information of velocity, x-direction and y-direction acceleration. The fused feature \( F_{\text{fuse}} \in \mathbb{R}^{(C'_b+3) \times L} \) is then fed into a Transformer encoder for enhanced representation.

\textbf{Control Prediction}: A Transformer decoder predicts control signals in an auto-regressive method. The fused feature \( F_{\text{fuse}} \) serves as keys \( K \) and values \( V \) in cross-attention, while the empty sequence serves as queries \( Q \). The decoder iteratively generates control tokens, which are detokenized into normalized acceleration, normalized steering angle and gear state, then mapped to control commands.

\subsection{Loss}

There are two main loss functions and one auxiliary task:

\textbf{Control Signal Loss}: A cross-entropy loss is used to minimize the discrepancy between predicted and ground truth control sequences.

\textbf{Semantic Segmentation Loss}: The model classifies BEV grid objects into \textit{vehicle}, \textit{target slot}, and \textit{background}, supervised with cross-entropy loss.

\textbf{Auxiliary Task}: Inspired by BEVDepth \cite{li2023bevdepth}, depth prediction is applied to enhance interpretability and improve performance.

\section{Dataset Generation}
\subsection{Implementation Details}\label{Implementation}
\begin{figure}
    \centering
    \includegraphics[width=1\linewidth]{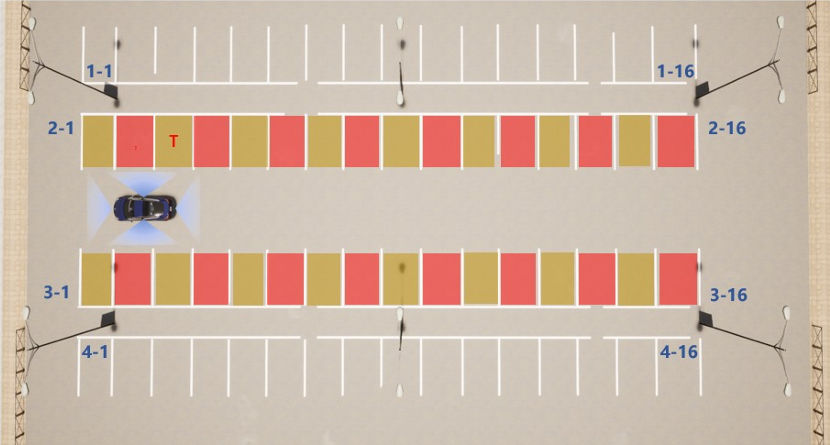}
        \caption{Top view of the parking lot in CARLA}
    \label{fig:parking_lot}
\end{figure}

The simulator is CARLA 0.9.11, which keeps the same as the one in the original paper. Figure~\ref{fig:parking_lot} shows the distribution of slots in the parking lot, where the red letter ``T'' designates the target slot. The red parking slots 2-2, ..., 2-16, 3-2, ..., 3-16 which end with even number are used for data generation and the brown parking slots 2-1, ..., 2-15, 3-1, ..., 3-15 which end with odd number are used for closed-loop test. Most of the settings are the same as the original ones. The ego vehicle is initialized at a random position of the path between the second and the third row, heading parallel with the path.

\subsection{Problems and Solutions}\label{Problems}

Through preliminary reproduction of dataset and corresponding tests, there exist following problems. The core of dataset optimization is to solve these problems, which is the guideline of subsequent experiments.

\subsubsection{Keyboard Control}

The original authors used a keyboard for vehicle control with four basic operations: throttle, brake, steer, and gear shift.

\begin{itemize}
    \item \textbf{Throttle}: Press \textbf{W} or \textbf{Up Arrow} to increase throttle by 0.05 per frame (up to a maximum of 0.5). Releasing resets to 0.
    \item \textbf{Brake}: Press \textbf{S}, \textbf{Down Arrow}, or \textbf{Space} to increase brake intensity by 0.2 per frame (up to a maximum of 1). Releasing resets to 0.
    \item \textbf{Steering}: Press \textbf{A/Left Arrow} or \textbf{D/Right Arrow} to adjust steering based on frame time:
    $ \text{steer\_cache} \pm= 5 \times 10^{-4} \times \text{milliseconds} $.
    Steering angle is clamped between -0.7 and 0.7, rounded to one decimal place. Releasing resets to 0. If a steering key is pressed in the opposite direction, \text{steer\_cache} resets to 0 before applying the new input.
    \item \textbf{Gear Shift}: Press \textbf{Q} to toggle between forward (1) and reverse (-1).
\end{itemize}

Keyboard control is challenging for parking due to its discrete nature, which does not intuitively correspond to the continuous control inputs required for smooth maneuvering. To improve control, an XBOX controller was implemented to allow direct linear input. Triggers act as accelerator/brake pedals, and joysticks resemble a steering wheel (self-centering when released). This provides smoother, more intuitive control while maintaining the original value ranges and resolutions.

\subsubsection{Insufficient scenes and dataset size}
Analysis of the provided code indicates that the \texttt{--random\_seed} argument in the \texttt{argparser} lacks an explicit \texttt{int} type definition. As a result, the specified random seed value was interpreted as a string, leading to an error when invoked. This suggests that the authors used only the default random seed value (0), with its increment tied to the task index. Consequently, each parking slot was assigned only one single random seed for scene creation, limiting scene variation.

The original authors use random seeds from 0 to 15 for the 16 parking slots. As an improvement, we used random seeds from 0 to 15, assigning each to one parking slot and collecting eight parking routes per slot. The process was then repeated with seeds from 16 to 31 for another 16 slots. Random seeds 32–47 were also assigned to 16 slots, but only four routes were collected per slot. 

\subsubsection{Lack of parking slot with shadow of streetlight}

Our experiments demonstrated that the trained model fails to park the vehicle into a parking slot with streetlight shadows when strictly following the data generation pipeline. This issue arises because only slot 3-9 contain shadows, which are used for model test. During data sampling, the vehicle was never parked into a slot with a shadow, leading to difficulties during testing with the slot 3-9.

To capture parking slots with shadows while adhering to the original authors' division of training and testing slots, we decided to adjust the solar altitude and azimuth angles. Two scenarios were generated, with the shadow projected onto slots 3-8 and 3-10, respectively. When a shadow is projected onto slot 3-8, random seeds 11 and 27 were assigned to collect 8 routes per seed. Similarly, when a shadow is projected onto slot 3-10, random seeds 12 and 28 were used to collect 8 routes per seed.

\subsubsection{Trouble of parking slots near the edge of the parking lot}

During testing, when the target parking slots are 2-1 and 3-1 and the vehicle initially faces outward from the parking lot, the ideal maneuver involves driving straight to the edge of the parking lot, then adjusting the steering and reversing into the slot. However, in the training set, parking routes were collected for neighboring slots 2-2 and 3-2, without instances of the vehicle reaching the edge of the parking lot. As a result, during testing, the model encounters previously unseen scenarios when handling slots 2-1 and 3-1, leading to reduced stability.

As a solution, compared to the original speed limits of 12 km/h and 10 km/h for forwarding and backing, the speed is now further limited to 7.5 km/h for both directions. Experiments have proven that this adjustment improves the Target Success Rate.

\subsection{Dataset Overview}

Each route data of the dataset was generated via the script provided by original authors. The sampling frequency is 10 \text{Hz}. The data for model input is RGB images from 4 cameras, extrinsic and intrinsic matrix for each camera, target point and motion state of ego vehicle, while control values, depth and segmentation are ground-truth labels. The range and resolution of the control values are listed in Table~\ref{Range and resolution}. 5 versions of dataset were created for comparison experiments by one expert driver: Gen 1A, 1B, 1C, 2A, 2B.

\begin{table}[h]
\caption{Range and resolution of control values}\label{Range and resolution}%
\renewcommand\arraystretch{1.2}
\begin{tabular}{@{\hspace{0pt}}m{2.8cm}<{\centering}@{\hspace{0pt}}@{\hspace{0pt}}m{3cm}<{\centering}@{\hspace{0pt}}@{\hspace{0pt}}m{2.8cm}<{\centering}}
\toprule
Control & Range     & Resolution \\ 
\midrule
Throttle & [0, 0.5]  & 0.1 \\
Brake    & [0, 1]    & 0.2 \\
Steering    & [-0.7, 0.7] & 0.1 \\
Reverse  & \{0, 1\}    & --  \\ 
\bottomrule
\end{tabular}
\end{table}

\begin{figure}[ht]
    \centering
    \begin{subfigure}[b]{0.31\linewidth}
        \centering
        \includegraphics[width=\linewidth]{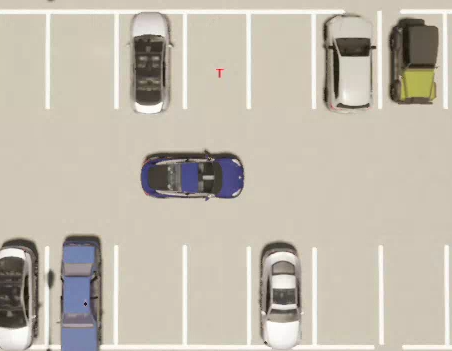}
        \caption{Initialization}
        \label{fig:Gen1f0}
    \end{subfigure}
    \hfill
    \begin{subfigure}[b]{0.31\linewidth}
        \centering
        \includegraphics[width=\linewidth]{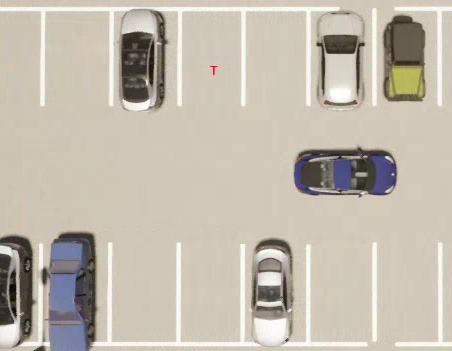}
        \caption{Stop}
        \label{fig:Gen1f1}
    \end{subfigure}
    \hfill
    \begin{subfigure}[b]{0.31\linewidth}
        \centering
        \includegraphics[width=\linewidth]{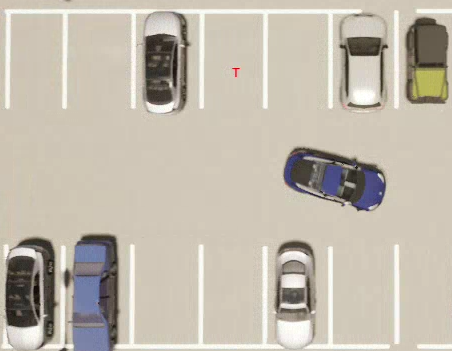}
        \caption{Reverse and steer}
        \label{fig:Gen1f2}
    \end{subfigure}

    \vspace{0.5cm}

    \begin{subfigure}[b]{0.31\linewidth}
        \centering
        \includegraphics[width=\linewidth]{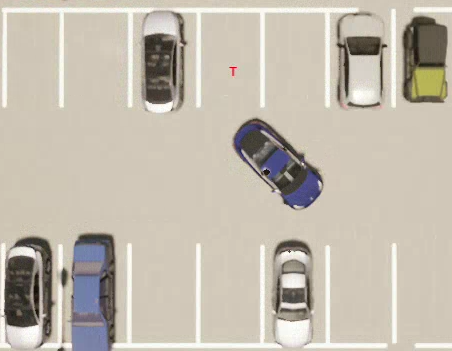}
        \caption{Reverse and steer}
        \label{fig:Gen1f3}
    \end{subfigure}
    \hfill
    \begin{subfigure}[b]{0.31\linewidth}
        \centering
        \includegraphics[width=\linewidth]{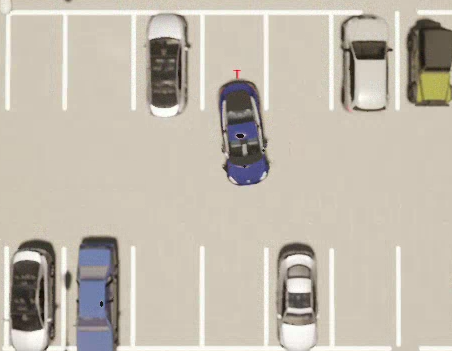}
        \caption{Fine tuning}
        \label{fig:Gen1f4}
    \end{subfigure}
    \hfill
    \begin{subfigure}[b]{0.31\linewidth}
        \centering
        \includegraphics[width=\linewidth]{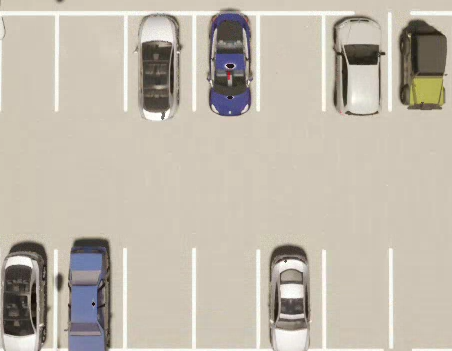}
        \caption{End}
        \label{fig:Gen1f5}
    \end{subfigure}
    
    \caption{Parking strategy in Gen 1}
    \label{fig:Parking strategy in Gen 1}
\end{figure}

In Gen 1, after driving straight, the vehicle stops at a position slightly farther from the target parking slot, as is shown in Figure~\ref{fig:Gen1f1}. It then gradually increases the steering angle while reversing. When both rear wheels cross the extended boundary line of the target slot (like Figure~\ref{fig:Gen1f3}), the steering angle is increased to its maximum. Subsequently, the steering and throttle values are gradually reduced. As the vehicle's front approaches alignment, the steering angle is fine-tuned (like Figure~\ref{fig:Gen1f4}). Once aligned, the throttle is further adjusted in reverse until the vehicle reaches the desired parking position (Figure~\ref{fig:Gen1f5}), at which point the brake is fully applied. During the model test for parking slot 2-1 and 3-1, the initialization of the vehicle is outside the parking lot, with half of the body on the road. Therefore, during the data generation for 2-2 and 3-2, original initialization was modified to make the vehicle generated at that position in 10 routes, so that the model would have potential to handle this type of situation.

\begin{figure}[ht]
    \centering
    \begin{subfigure}[b]{0.31\linewidth}
        \centering
        \includegraphics[width=\linewidth]{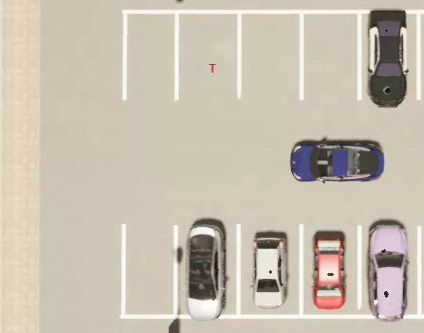}
        \caption{Initialization}
        \label{fig:Gen2f0}
    \end{subfigure}
    \hfill
    \begin{subfigure}[b]{0.31\linewidth}
        \centering
        \includegraphics[width=\linewidth]{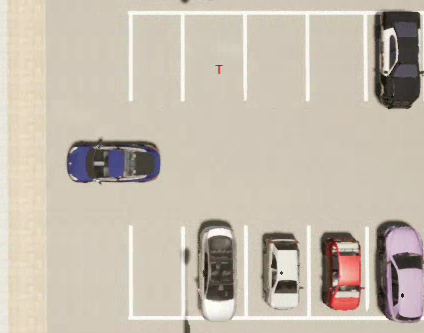}
        \caption{Stop}
        \label{fig:Gen2f1}
    \end{subfigure}
    \hfill
    \begin{subfigure}[b]{0.31\linewidth}
        \centering
        \includegraphics[width=\linewidth]{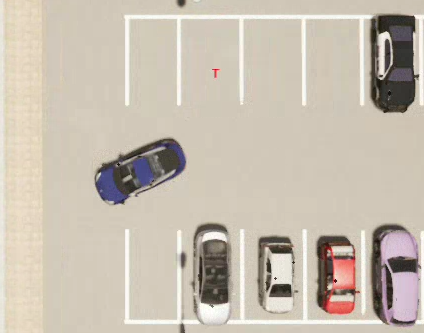}
        \caption{Reverse and steer}
        \label{fig:Gen2f2}
    \end{subfigure}

    \vspace{0.5cm}

    \begin{subfigure}[b]{0.31\linewidth}
        \centering
        \includegraphics[width=\linewidth]{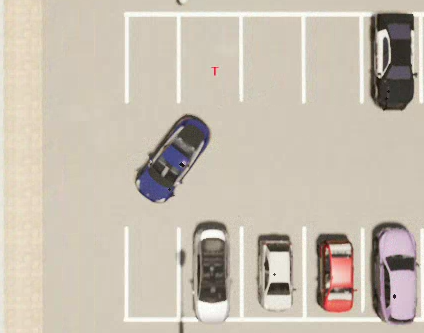}
        \caption{Reverse and steer}
        \label{fig:Gen2f3}
    \end{subfigure}
    \hfill
    \begin{subfigure}[b]{0.31\linewidth}
        \centering
        \includegraphics[width=\linewidth]{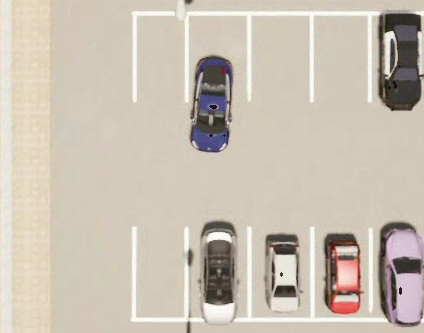}
        \caption{Fine tuning}
        \label{fig:Gen2f4}
    \end{subfigure}
    \hfill
    \begin{subfigure}[b]{0.31\linewidth}
        \centering
        \includegraphics[width=\linewidth]{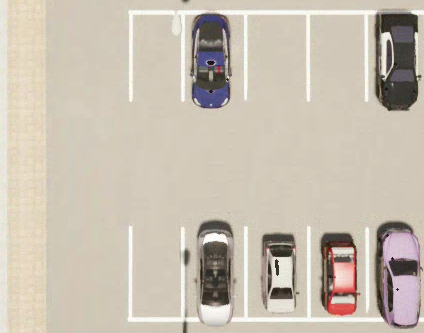}
        \caption{End}
        \label{fig:Gen2f5}
    \end{subfigure}
    
    \caption{Parking strategy in Gen 2}
    \label{fig:Parking strategy in Gen 2}
\end{figure}

In Gen 2, the vehicle stops closer to the target parking slot after driving straight compared to Gen 1, as Figure~\ref{fig:Gen2f1} shows. It then reverses with the steering angle at its maximum. As the rear approaches the slot, the steering and throttle values are gradually decreased (like Figure~\ref{fig:Gen2f2} and \ref{fig:Gen2f3}). The subsequent operations are the same as those in Gen 1. There is a slight change for initialization compared to Gen 1, in order to remove the instability caused by roadside shoulder. In Gen 2, the initial positions of parking slots 2-1 and 3-1 during testing are always within the parking lot, whereas for some routes of slots 2-2 and 3-2 during data generation, the vehicle is initialized near the edge of the parking lot to ensure coverage of edge-adjacent scenarios.

The selection of random seeds for each dataset is listed in~\ref{tab:trajectories}. The random seed ranges refer to the seeds employed in the scene generation process, which can change the total number of vehicles and the type of vehicle in each parking slot.




\section{Experiments}\label{Experiments}

\begin{table*}[h]
\caption{Number of trajectories in each dataset with different random seeds}
\renewcommand\arraystretch{1}
\small
\begin{minipage}{\textwidth}
\renewcommand{\thefootnote}{\arabic{footnote}}
\setcounter{footnote}{0}
\centering
\begin{tabular}{@{\hspace{0pt}}m{7cm}<{\centering}@{\hspace{0pt}}@{\hspace{0pt}}m{2.2cm}<{\centering}@{\hspace{0pt}}@{\hspace{0pt}}m{2.2cm}<{\centering}@{\hspace{0pt}}@{\hspace{0pt}}m{2.2cm}<{\centering}@{\hspace{0pt}}@{\hspace{0pt}}m{2.2cm}<{\centering}@{\hspace{0pt}}@{\hspace{0pt}}m{2.2cm}<{\centering}@{\hspace{0pt}}@{\hspace{0pt}}m{2.2cm}<{\centering}@{\hspace{0pt}}}
\toprule
Scenario & \multicolumn{5}{c}{Trajectories (Train / Val) in Dataset} \\
\cmidrule(lr){2-6}
 & Gen 1A & Gen 1B & Gen 1C & Gen 2A & Gen 2B \\
\midrule
Seed 0 to 15 & 96 / 32 & 96 / 32 & 96 / 32 & 96 / 32 & 96 / 32 \\
Seed 16 to 31 &  & 96 / 32 & 96 / 32 &  & 96 / 32 \\
Seed 32 to 47 &  &  &  &  & 32 / 32 \\
Seed 16 to 143 &  &  &  & 96 / 32\footnote{Gen 2A has more scenes. For example, random seeds 16, 32, 48, 64, 80, 96, 112 and 128 are assigned to the first parking lot for generation.} &  \\
3-8 (shadow), seed 11, 27 &  &  & 12 / 4 &  & 12 / 4 \\
3-10 (shadow), seed 12, 28 &  &  & 12 / 4\footnote{Additionally, in Gen 1C, routes are generated for slot 3-8 and 3-10 with shadow separately after modifying the sun azimuth angle and altitude angle. For example, when random seed 11 is assigned to slot 3-8 in scene generation, 6 routes are for training set and 2 routes are for validation set. The same for seed 27.} &  & 12 / 4 \\
3-8 (shadow), seed 91, 107, 123, 139 &  &  &  & (3 / 1) & \\
3-10 (shadow), seed 28, 44, 60, 76 &  &  &  & (3 / 1)\footnote{For slot 3-8 and 3-10 in Gen 2A, the routes with shadow in the slots are not generated separately, but are included in the cases with random seeds ranging from 16 to 143 mentioned above. Therefore, these routes are not counted in the total.} & (1 / 1)\footnote{The routes for training and validation are assigned seeds 28 and 44, respectively, both of which are included in the cases mentioned above.} \\
\midrule
Total & 96 / 32 & 192 / 64 & 216 / 72 & 192 / 64 & 248 / 104 \\
\bottomrule
\end{tabular}
\end{minipage}
\label{tab:trajectories}
\end{table*}

\subsection{Experimental Setup}\label{setup}

The network architecture keeps the same as the original one and hyperparameters are slightly different. The network was trained on 4 NVIDIA RTX 3090 (total VRAM: 96 GB) for dataset Gen 1A, 1B, 1C and 2A, and on 4 NVIDIA RTX 4090 (total VRAM: 96 GB) for dataset Gen 2B, both with a batch size of 16. Adam optimizer is utilized with a weight decay of $1 \times 10^{-4}$, while the beta values are 0.9 and 0.999 as default. The performance of models trained on 5 datasets with different epochs and initial learning rate will be compared.

\subsubsection{Iteration of datasets}\label{datasets}

Gen 1A has 128 routes with random scenes, with 96 for training and 32 for validation. Each parking slot has only 1 random seed to generate a scene. Random seeds 0 to 15 are allocated to 16 parking slots in order. For each parking slot 8 routes are collected, where 6 routes are designated for the training set, and 2 routes are designated for the validation set. The initial positions of the 6 training routes are distributed as far-left, middle-left, near-left, near-right, middle-right, and far-right from a bird's-eye view. For the 2 validation routes, their initial positions are relatively random, with the constraint that one is positioned on the left and the other on the right.


Gen 1B inherited 128 trajectories from Gen 1A and introduced an additional 128 trajectories, with random seeds 16–31 used for scene generation. The settings of initial positions and the split of training and validation set remain the same.


Building upon Gen 1B, Gen 1C further incorporated 16 trajectories for slot 3-8 with shadow and 16 trajectories for slot 3-10 with shadow. 24 of them are for training and 8 of them are for validation. The initial position settings and the division of the training and validation sets remain unchanged.


Similar to Gen 1A, 128 trajectories were generated for Gen 2A using random seeds 0 to 15. The difference is that the straight-line distance before reversing was shortened to prevent the vehicle from driving out of the parking lot. To enhance the diversity, more random seeds are used to generate scenes for the extra 128 routes, as~\ref{tab:weather_settings} illustrates. Each random seed is assigned to one of 16 parking slots with different weather settings to ensure a diverse dataset. Weather 0 is default weather, in which the shadow of streetlight is projected on slot 3-9. In Weather 1, the sun angles are set to project the shadow on slot 3-10, while in Weather 2, the shadow is on slot 3-8.

\begin{table*}[h]
\caption{Random seed and initial position assignment w.r.t weather settings in training and validation sets in Gen 2A}
\renewcommand\arraystretch{1.5}
\label{tab:weather_settings}
\begin{tabular}{@{\hspace{0pt}}m{4.5cm}<{\centering}@{\hspace{0pt}}@{\hspace{0pt}}m{4.5cm}<{\centering}@{\hspace{0pt}}@{\hspace{0pt}}m{4.5cm}<{\centering}@{\hspace{0pt}}@{\hspace{0pt}}m{4.5cm}<{\centering}@{\hspace{0pt}}}
\toprule
Dataset & Weather 0 & Weather 1 & Weather 2 \\
\midrule
\multirow{3}{*}{Training set} & \multirow{3}{*}{6 initial positions: 0 to 15} & far-left: 16 to 31 & near-right: 80 to 95 \\
 &  & middle-left: 32 to 47 & middle-right: 96 to 111 \\
 &  & near-left: 48 to 63 & far-right: 112 to 127 \\
\midrule
Validation set & flexible: 0 to 15 & flexible: 64 to 79 & flexible: 128 to 143 \\
\bottomrule
\end{tabular}
\end{table*}


Compared to Gen 2A, Gen 2B eliminates the selection of random seeds ranging from 48 to 143. It retains 160 routes from Gen 2A, comprising 128 routes associated with seeds 0 to 15, 16 routes with seeds 16 to 31, and 16 routes with seeds 32 to 47. Including these inherited routes, there are 128 routes assigned to seeds 16 to 31, while 64 routes are assigned to seeds 32 to 47. For these 64 routes, each slot is allocated 2 routes for the training set and 2 routes for the validation set. The initial positions of the 2 training routes are set to far-left and far-right, whereas those of the 2 validation routes are positioned at middle-left and middle-right. Furthermore, 32 additional routes are generated for slots 3-8 and 3-10 with shadows, following the same initialization distribution and random seed selection strategy as the corresponding part in Gen 1C.

\subsection{Metrics}\label{Metrics}

The metrics used for model testing remain the same as the original ones.

\textbf{Target Success Rate (TSR)} represents the probability of the ego vehicle successfully parking in the designated slot. A parking attempt is considered successful if the vehicle's center is within 0.6 meters horizontally and 1 meter longitudinally from the slot's center, with an orientation deviation of no more than 10 degrees.

\textbf{Target Failure Rate (TFR)} refers to the probability of the ego vehicle reaching the target parking slot but with errors exceeding acceptable limits.

\textbf{Non-Target Rate (NTR)} represents the probability that the ego vehicle parks in a slot other than the designated target. \textbf{NTR} is the sum of the \textbf{Non-Target Success Rate (NTSR)} and \textbf{Non-Target Failure Rate (NTFR)}, both of which depend on whether the error in parking into a non-target slot is acceptable.

\textbf{Collision Rate (CR)} refers to the probability of a collision occurring during the parking process.

\textbf{Outbound Rate (OR)} indicates the probability that the ego vehicle exits the parking area.

\textbf{Timeout Rate (TR)} refers to the probability that the ego vehicle fails to park successfully within a specified time limit. The original authors combine \textbf{OR} and \textbf{TR} and refers to their sum as \textbf{TR}.

\textbf{Average Position Error (APE)} is the mean error between the ego vehicle's final position and the center of the target slot in successful parking instances.

\textbf{Average Orientation Error (AOE)} is the average difference between the ego vehicle's final yaw angle and the desired orientation of the target slot in successful parking instances.

\textbf{Average Parking Time (APT)} represents the mean duration taken for successful parking maneuvers.

\textbf{Average Inference Time (AIT)} is the average time spent on network inference per step.

\subsection{Results}\label{Results}

To test the model efficiently, each model is required to accomplish 96 parking tasks (1 test epoch), with 6 tasks assigned to each of the 16 test parking slots, each task having a different initial position for the vehicle. The best model trained on Gen 2B will be tested for 4 test epochs with 384 tasks to ensure the stability of the test.

\subsubsection{Details of Gen 1A testing}

According to Table~\ref{tab:Gen_1A}, the model trained on the Gen 1A dataset performed poorly in testing, achieving a TSR of only 44.792\% after 155 epochs of training. For comparison, the model trained on the Gen 1B dataset achieved a TSR of 65.625\% in testing after 155 epochs with the same learning rate of $1 \times 10^{-4}$, benefiting from the doubled dataset size and increased scenario diversity.

\begin{table}[h]
\caption{Gen 1A's Performance}\label{tab:Gen_1A}%
\renewcommand\arraystretch{1.4}
\centering
\begin{tabular}{@{\hspace{0pt}}m{3cm}<{\centering}@{\hspace{0pt}}@{\hspace{0pt}}m{2.8cm}<{\centering}@{\hspace{0pt}}@{\hspace{0pt}}m{2.8cm}<{\centering}@{\hspace{0pt}}}
\toprule
\textbf{Metric} & \multicolumn{2}{c}{\textbf{Lr = $1 \times 10^{-4}$}} \\
\midrule
Epochs         & 120    & 155    \\
val\_loss & 0.713 & 0.706 \\
TSR(\%) & 40.625 & 44.792 \\
\bottomrule
\end{tabular}
\end{table}

Table~\ref{tab:Gen_1A_default} presents the test results of the best model trained on Gen 1A. Each of the 16 parking slot is tested for 6 times with different initial positions. For parking slots 2-1, 2-3, 3-1, and 3-9, the model completely failed to park the vehicle into the target slot. For slots 2-1 and 3-1, the main problem is the high OR, which means the vehicle frequently exits from the parking lot abnormally. For slot 2-3, the vehicle collides with the neighboring vehicle very often. For slot 3-9, the model has a misunderstanding with the shadow on the ground and refuses to park the vehicle into the target slot, which causes timeout and collision. For other parking slots, the model can't ensure a robust performance and the TSR varies from slot to slot.

\begin{table*}[htbp]
\small
\caption{Closed-loop results including 96 test cases for the best model trained on Gen 1A}\label{tab:Gen_1A_default}
\renewcommand\arraystretch{1.5}
\centering
\begin{tabular}{@{\hspace{0pt}}m{1.5cm}<{\centering}@{\hspace{0pt}}@{\hspace{0pt}}m{1.5cm}<{\centering}@{\hspace{0pt}}@{\hspace{0pt}}m{1.5cm}<{\centering}@{\hspace{0pt}}@{\hspace{0pt}}m{1.5cm}<{\centering}@{\hspace{0pt}}@{\hspace{0pt}}m{1.5cm}<{\centering}@{\hspace{0pt}}@{\hspace{0pt}}m{1.5cm}<{\centering}@{\hspace{0pt}}@{\hspace{0pt}}m{1.5cm}<{\centering}@{\hspace{0pt}}@{\hspace{0pt}}m{1.5cm}<{\centering}@{\hspace{0pt}}@{\hspace{0pt}}m{1.5cm}<{\centering}@{\hspace{0pt}}@{\hspace{0pt}}m{1.5cm}<{\centering}@{\hspace{0pt}}@{\hspace{0pt}}m{1.5cm}<{\centering}@{\hspace{0pt}}@{\hspace{0pt}}m{1.5cm}<{\centering}@{\hspace{0pt}}}
\toprule
TaskIdx & TSR (\%) & TFR (\%) & NTSR (\%) & NTFR (\%) & CR (\%) & OR (\%) & TR (\%) & APE (m) & AOE (deg) & APT (s) & AIT (s) \\
\midrule
2-1  & 0.000 & 0.000 & 16.667 & 0.000 & 16.667 & 50.000 & 16.667 & -- & -- & -- & 0.080 \\
2-3  & 0.000 & 0.000 & 16.667 & 0.000 & 66.667 & 0.000  & 16.667 & -- & -- & -- & 0.082 \\
2-5  & 50.000 & 0.000 & 33.333 & 0.000 & 0.000 & 0.000 & 16.667 & 0.483 & 0.696 & 21.667 & 0.081 \\
2-7  & 66.667 & 0.000 & 0.000 & 0.000 & 33.333 & 0.000 & 0.000 & 0.486 & 0.686 & 22.000 & 0.082 \\
2-9  & 83.333 & 0.000 & 0.000 & 0.000 & 0.000 & 0.000 & 16.667 & 0.555 & 1.154 & 23.873 & 0.081 \\
2-11 & 50.000 & 0.000 & 0.000 & 0.000 & 50.000 & 0.000 & 0.000 & 0.566 & 1.868 & 22.056 & 0.084 \\
2-13 & 50.000 & 16.667 & 0.000 & 0.000 & 33.333 & 0.000 & 0.000 & 0.476 & 0.921 & 22.289 & 0.087 \\
2-15 & 100.000 & 0.000 & 0.000 & 0.000 & 0.000 & 0.000 & 0.000 & 0.387 & 0.951 & 22.178 & 0.079 \\
3-1  & 0.000 & 0.000 & 0.000 & 0.000 & 16.667 & 66.667 & 16.667 & -- & -- & -- & 0.079 \\
3-3  & 100.000 & 0.000 & 0.000 & 0.000 & 0.000 & 0.000 & 0.000 & 0.387 & 1.244 & 23.961 & 0.080 \\
3-5  & 33.333 & 0.000 & 0.000 & 0.000 & 0.000 & 50.000 & 16.667 & 0.307 & 1.087 & 24.550 & 0.080 \\
3-7  & 66.667 & 0.000 & 16.667 & 0.000 & 16.667 & 0.000 & 0.000 & 0.446 & 1.515 & 21.775 & 0.080 \\
3-9  & 0.000 & 0.000 & 0.000 & 0.000 & 33.333 & 0.000 & 66.667 & -- & -- & -- & 0.080 \\
3-11 & 50.000 & 0.000 & 16.667 & 0.000 & 0.000 & 33.333 & 0.000 & 0.445 & 1.738 & 22.456 & 0.079 \\
3-13 & 16.667 & 0.000 & 0.000 & 0.000 & 0.000 & 33.333 & 50.000 & 0.348 & 0.903 & 24.233 & 0.079 \\
3-15 & 50.000 & 0.000 & 0.000 & 0.000 & 0.000 & 16.667 & 33.333 & 0.285 & 0.279 & 24.133 & 0.079 \\
\textbf{Avg}  & \textbf{44.792} & \textbf{1.042} & \textbf{6.250} & \textbf{0.000} & \textbf{16.667} & \textbf{15.625} & \textbf{15.625} & \textbf{0.431} & \textbf{1.087} & \textbf{22.931} & \textbf{0.081} \\
\bottomrule
\end{tabular}
\end{table*}

\subsubsection{Details of Gen 1B testing}

A comparative experiment on different initial learning rates was conducted using the Gen 1B dataset. An initial learning rate of $5 \times 10^{-4}$ caused oscillations in the training loss, so no testing was performed. Meanwhile, an initial learning rate of $5 \times 10^{-5}$ results in a TSR of only around 50\%, which indicates an appropriate learning rate should be chosen between these two values. The model trained with initial learning rate $Lr = 7.5 \times 10^{-5}$ after 195 epochs has the highest TSR of 71.875\%.

\begin{table*}[htbp]
\caption{Comparison of GEN 1B's Performance under Different Initial Learning Rates}
\renewcommand\arraystretch{1.5}
\label{tab:Gen_1B_comparison}
\centering
\small
\begin{tabular}{@{\hspace{0pt}}m{1.8cm}<{\centering}@{\hspace{0pt}}@{\hspace{0pt}}m{1.8cm}<{\centering}@{\hspace{0pt}}@{\hspace{0pt}}m{1.8cm}<{\centering}@{\hspace{0pt}}@{\hspace{0pt}}m{1.8cm}<{\centering}@{\hspace{0pt}}@{\hspace{0pt}}m{1.8cm}<{\centering}@{\hspace{0pt}}@{\hspace{0pt}}m{1.8cm}<{\centering}@{\hspace{0pt}}@{\hspace{0pt}}m{1.8cm}<{\centering}@{\hspace{0pt}}@{\hspace{0pt}}m{1.8cm}<{\centering}@{\hspace{0pt}}@{\hspace{0pt}}m{1.8cm}<{\centering}@{\hspace{0pt}}@{\hspace{0pt}}m{1.8cm}<{\centering}@{\hspace{0pt}}}
\toprule
\multirow{2}{*}{\textbf{Metric}} & \multicolumn{3}{c}{\textbf{Lr = $1 \times 10^{-4}$}} & \multicolumn{3}{c}{\textbf{Lr = $7.5 \times 10^{-5}$}} & \multicolumn{3}{c}{\textbf{Lr = $5 \times 10^{-5}$}} \\
\cmidrule(lr){2-4} \cmidrule(lr){5-7} \cmidrule(lr){8-10}
 & Epochs & val\_loss & TSR(\%) & Epochs & val\_loss & TSR(\%) & Epochs & val\_loss & TSR(\%) \\
\midrule
\textbf{Early} & 115 & 0.737 & 57.292 & -- & -- & -- & -- & -- & -- \\
\textbf{Mid} & 155 & 0.725 & 65.625 & 155 & 0.700 & 65.625  & 155 & 0.721 & 50.000 \\
\textbf{Late} & --  & -- & -- & 195 & 0.695 & 71.875 & 200 & 0.720 & 51.042 \\
\bottomrule
\end{tabular}
\end{table*}

Table~\ref{tab:Gen_1B_default} depicts the detailed test results of the best model trained on Gen 1B. With the help of a larger dataset, the model demonstrates a partial ability to handle parking tasks at target slots located at the edge of the parking area. For slot 3-9 with shadow on the ground, the model still can't deal with the task at all. However, the TSRs for other parking slots have improved to varying degrees, with the overall average TSR increasing by 27.083 percentage points compared to Gen 1A, demonstrating the great potential of the larger size and more scenes of dataset.

\begin{table*}[htbp]
\small
\caption{Closed-loop results including 96 test cases for the best model trained on Gen 1B}\label{tab:Gen_1B_default}
\renewcommand\arraystretch{1.5}
\centering
\begin{tabular}{@{\hspace{0pt}}m{1.5cm}<{\centering}@{\hspace{0pt}}@{\hspace{0pt}}m{1.5cm}<{\centering}@{\hspace{0pt}}@{\hspace{0pt}}m{1.5cm}<{\centering}@{\hspace{0pt}}@{\hspace{0pt}}m{1.5cm}<{\centering}@{\hspace{0pt}}@{\hspace{0pt}}m{1.5cm}<{\centering}@{\hspace{0pt}}@{\hspace{0pt}}m{1.5cm}<{\centering}@{\hspace{0pt}}@{\hspace{0pt}}m{1.5cm}<{\centering}@{\hspace{0pt}}@{\hspace{0pt}}m{1.5cm}<{\centering}@{\hspace{0pt}}@{\hspace{0pt}}m{1.5cm}<{\centering}@{\hspace{0pt}}@{\hspace{0pt}}m{1.5cm}<{\centering}@{\hspace{0pt}}@{\hspace{0pt}}m{1.5cm}<{\centering}@{\hspace{0pt}}@{\hspace{0pt}}m{1.5cm}<{\centering}@{\hspace{0pt}}}
\toprule
TaskIdx & TSR (\%) & TFR (\%) & NTSR (\%) & NTFR (\%) & CR (\%) & OR (\%) & TR (\%) & APE (m) & AOE (deg) & APT (s) & AIT (s) \\
\midrule
2-1   & 33.333  & 0.000  & 0.000  & 0.000  & 0.000  & 66.667  & 0.000  & 0.409 & 0.295 & 23.100 & 0.078 \\
2-3   & 100.000 & 0.000  & 0.000  & 0.000  & 0.000  & 0.000   & 0.000  & 0.492 & 0.685 & 22.150 & 0.076 \\
2-5   & 83.333  & 16.667 & 0.000  & 0.000  & 0.000  & 0.000   & 0.000  & 0.415 & 0.446 & 22.313 & 0.078 \\
2-7   & 100.000 & 0.000  & 0.000  & 0.000  & 0.000  & 0.000   & 0.000  & 0.484 & 0.276 & 23.433 & 0.080 \\
2-9   & 83.333  & 0.000  & 0.000  & 0.000  & 0.000  & 0.000   & 16.667 & 0.445 & 0.493 & 23.653 & 0.078 \\
2-11  & 100.000 & 0.000  & 0.000  & 0.000  & 0.000  & 0.000   & 0.000  & 0.458 & 0.718 & 21.239 & 0.080 \\
2-13  & 66.667  & 0.000  & 33.333 & 0.000  & 0.000  & 0.000   & 0.000  & 0.436 & 0.231 & 22.983 & 0.114 \\
2-15  & 83.333  & 0.000  & 16.667 & 0.000  & 0.000  & 0.000   & 0.000  & 0.306 & 0.314 & 22.927 & 0.077 \\
3-1   & 16.667  & 0.000  & 0.000  & 0.000  & 0.000  & 83.333  & 0.000  & 0.217 & 0.270 & 24.600 & 0.076 \\
3-3   & 83.333  & 0.000  & 0.000  & 0.000  & 16.667 & 0.000   & 0.000  & 0.245 & 0.389 & 22.047 & 0.077 \\
3-5   & 83.333  & 0.000  & 0.000  & 0.000  & 0.000  & 0.000   & 16.667 & 0.296 & 0.630 & 24.007 & 0.077 \\
3-7   & 83.333  & 0.000  & 0.000  & 0.000  & 0.000  & 0.000   & 16.667 & 0.327 & 0.753 & 22.527 & 0.077 \\
3-9   & 0.000   & 0.000  & 33.333 & 0.000  & 0.000  & 0.000   & 66.667 & --   & --   & --    & 0.077 \\
3-11  & 66.667  & 0.000  & 0.000  & 0.000  & 0.000  & 0.000   & 33.333 & 0.405 & 0.447 & 24.517 & 0.077 \\
3-13  & 66.667  & 0.000  & 0.000  & 0.000  & 0.000  & 0.000   & 33.333 & 0.181 & 0.248 & 25.217 & 0.077 \\
3-15  & 100.000 & 0.000  & 0.000  & 0.000  & 0.000  & 0.000   & 0.000  & 0.209 & 0.396 & 23.206 & 0.077 \\
\textbf{Avg}   & \textbf{71.875} & \textbf{1.042} & \textbf{5.208} & \textbf{0.000} & \textbf{1.042} & \textbf{9.375} & \textbf{11.458} & \textbf{0.355} & \textbf{0.439} & \textbf{23.195} & \textbf{0.080} \\
\bottomrule
\end{tabular}
\end{table*}

\subsubsection{Details of Gen 1C testing}

As Table~\ref{tab:Gen_1C_comparison} demonstrates, with the additional parking trajectories in shadowed slots, Gen 1C achieved a higher TSR in shadowed parking slots and outperformed Gen 1B in overall TSR. Specifically, with an initial learning rate of $1 \times 10^{-4}$, the model trained on Gen 1C for 150 epochs achieved a TSR 7.25 percentage points higher than the model trained on Gen 1B for 155 epochs. Similarly, with an initial learning rate of $7.5 \times 10^{-5}$, the model trained on Gen 1C for 155 epochs outperformed the Gen 1B model trained for the same number of epochs by 11.46 percentage points.

In Gen 1C with an initial learning rate of $1 \times 10^{-4}$, the model made no progress in TSR between 150 and 190 training epochs. In contrast, with an initial learning rate of $7.5 \times 10^{-5}$, the TSR improved by 3.13 percentage points between training from 65 to 155 epochs, but there is no improvement after that.


Table~\ref{tab:Gen_1C_default} demonstrates the results of the best model trained on Gen 1C. With the help of the newly added parking trajectories featuring shadow projected on the target slot, the new model shows a significant improvement on slot 3-9, with the TSR increasing from 0 to 66.667\%. This improvement in slot 3-9 contributes to a 5.208 percentage point increase in the overall average TSR compared to Gen 1B. TSRs for other slots show no notable changes, and slots 2-1 and 3-1 still maintain very low TSR levels, which is the primary focus during creation of dataset Gen 2.

\begin{table*}[htbp]
\small
\caption{Closed-loop results including 96 test cases for the best model trained on Gen 1C}\label{tab:Gen_1C_default}
\renewcommand\arraystretch{1.5}
\centering
\begin{tabular}{@{\hspace{0pt}}m{1.5cm}<{\centering}@{\hspace{0pt}}@{\hspace{0pt}}m{1.5cm}<{\centering}@{\hspace{0pt}}@{\hspace{0pt}}m{1.5cm}<{\centering}@{\hspace{0pt}}@{\hspace{0pt}}m{1.5cm}<{\centering}@{\hspace{0pt}}@{\hspace{0pt}}m{1.5cm}<{\centering}@{\hspace{0pt}}@{\hspace{0pt}}m{1.5cm}<{\centering}@{\hspace{0pt}}@{\hspace{0pt}}m{1.5cm}<{\centering}@{\hspace{0pt}}@{\hspace{0pt}}m{1.5cm}<{\centering}@{\hspace{0pt}}@{\hspace{0pt}}m{1.5cm}<{\centering}@{\hspace{0pt}}@{\hspace{0pt}}m{1.5cm}<{\centering}@{\hspace{0pt}}@{\hspace{0pt}}m{1.5cm}<{\centering}@{\hspace{0pt}}@{\hspace{0pt}}m{1.5cm}<{\centering}@{\hspace{0pt}}}
\toprule
TaskIdx & TSR (\%) & TFR (\%) & NTSR (\%) & NTFR (\%) & CR (\%) & OR (\%) & TR (\%) & APE (m) & AOE (deg) & APT (s) & AIT (s) \\
\midrule
2-1   & 16.667  & 0.000  & 33.333  & 0.000  & 16.667  & 33.333  & 0.000  & 0.383 & 0.363 & 26.800 & 0.077 \\
2-3   & 83.333  & 0.000  & 0.000   & 0.000  & 0.000   & 0.000   & 16.667 & 0.304 & 0.258 & 22.887 & 0.077 \\
2-5   & 50.000  & 0.000  & 0.000   & 0.000  & 0.000   & 0.000   & 50.000 & 0.377 & 0.608 & 23.989 & 0.077 \\
2-7   & 100.000 & 0.000  & 0.000   & 0.000  & 0.000   & 0.000   & 0.000  & 0.352 & 0.284 & 23.061 & 0.077 \\
2-9   & 83.333  & 0.000  & 16.667  & 0.000  & 0.000   & 0.000   & 0.000  & 0.462 & 0.338 & 22.300 & 0.077 \\
2-11  & 100.000 & 0.000  & 0.000   & 0.000  & 0.000   & 0.000   & 0.000  & 0.410 & 0.464 & 22.811 & 0.078 \\
2-13  & 100.000 & 0.000  & 0.000   & 0.000  & 0.000   & 0.000   & 0.000  & 0.354 & 0.988 & 22.122 & 0.077 \\
2-15  & 100.000 & 0.000  & 0.000   & 0.000  & 0.000   & 0.000   & 0.000  & 0.345 & 1.093 & 23.206 & 0.077 \\
3-1   & 16.667  & 0.000  & 0.000   & 16.667 & 16.667  & 33.333  & 16.667 & 0.290 & 1.019 & 23.333 & 0.077 \\
3-3   & 66.667  & 0.000  & 16.667  & 0.000  & 16.667  & 0.000   & 0.000  & 0.240 & 0.451 & 24.283 & 0.077 \\
3-5   & 100.000 & 0.000  & 0.000   & 0.000  & 0.000   & 0.000   & 0.000  & 0.328 & 0.211 & 23.550 & 0.078 \\
3-7   & 100.000 & 0.000  & 0.000   & 0.000  & 0.000   & 0.000   & 0.000  & 0.344 & 0.707 & 21.356 & 0.078 \\
3-9   & 66.667  & 0.000  & 16.667  & 0.000  & 16.667  & 0.000   & 0.000  & 0.264 & 0.984 & 23.242 & 0.077 \\
3-11  & 66.667  & 0.000  & 33.333  & 0.000  & 0.000   & 0.000   & 0.000  & 0.240 & 0.361 & 23.775 & 0.077 \\
3-13  & 83.333  & 0.000  & 0.000   & 0.000  & 0.000   & 0.000   & 16.667 & 0.187 & 0.307 & 25.693 & 0.077 \\
3-15  & 100.000 & 0.000  & 0.000   & 0.000  & 0.000   & 0.000   & 0.000  & 0.176 & 0.339 & 24.506 & 0.077 \\
\textbf{Avg}   & \textbf{77.083} & \textbf{0.000} & \textbf{7.292} & \textbf{1.042} & \textbf{4.167} & \textbf{4.167} & \textbf{6.250} & \textbf{0.316} & \textbf{0.548} & \textbf{23.557} & \textbf{0.077} \\
\bottomrule
\end{tabular}
\end{table*}

\subsubsection{Details of Gen 2A testing}

Based on the experience gained from Gen 1, an initial learning rate of $7.5 \times 10^{-5}$ will be chosen for Gen 2, with a maximum of 155 epochs of training, to achieve a balance between model performance and training duration. The result is demonstrated in Table~\ref{tab:Gen_2A_2B}.

In Gen 2A, increasing the number of additional scenes without increasing the dataset size did not have a positive effect on model performance, as the TSR of 63.542\% was 13.54 percentage points lower than the best result achieved by Gen 1C.

Table~\ref{tab:Gen_2A_default} demonstrates the test results of the model trained on Gen 2A. Compared to Gen 1C, the new model performs better with slot 2-1, 2-5 and 3-1, which indicates that the shortened forward-driving distance before reversing is helpful in handling the scenarios near the edge of the parking lot. However, there are 10 slots where the TSR drops with the new model, which proves that t is not adequate to generate only one trajectory for a single parking slot in a given scenario and it is unwise to increase the number of parking slot scenarios while keeping the total number of trajectories unchanged. Overly complicated scenario setups are not an effective solution for optimizing the dataset.

\begin{table*}[htbp]
\small
\caption{Closed-loop results including 96 test cases for the model trained on Gen 2A}\label{tab:Gen_2A_default}
\renewcommand\arraystretch{1.5}
\centering
\begin{tabular}{@{\hspace{0pt}}m{1.5cm}<{\centering}@{\hspace{0pt}}@{\hspace{0pt}}m{1.5cm}<{\centering}@{\hspace{0pt}}@{\hspace{0pt}}m{1.5cm}<{\centering}@{\hspace{0pt}}@{\hspace{0pt}}m{1.5cm}<{\centering}@{\hspace{0pt}}@{\hspace{0pt}}m{1.5cm}<{\centering}@{\hspace{0pt}}@{\hspace{0pt}}m{1.5cm}<{\centering}@{\hspace{0pt}}@{\hspace{0pt}}m{1.5cm}<{\centering}@{\hspace{0pt}}@{\hspace{0pt}}m{1.5cm}<{\centering}@{\hspace{0pt}}@{\hspace{0pt}}m{1.5cm}<{\centering}@{\hspace{0pt}}@{\hspace{0pt}}m{1.5cm}<{\centering}@{\hspace{0pt}}@{\hspace{0pt}}m{1.5cm}<{\centering}@{\hspace{0pt}}@{\hspace{0pt}}m{1.5cm}<{\centering}@{\hspace{0pt}}}
\toprule
TaskIdx & TSR (\%) & TFR (\%) & NTSR (\%) & NTFR (\%) & CR (\%) & OR (\%) & TR (\%) & APE (m) & AOE (deg) & APT (s) & AIT (s) \\
\midrule
2-1   & 50.000  & 0.000  & 0.000  & 0.000  & 16.667  & 33.333  & 0.000  & 0.247 & 0.594 & 19.111 & 0.077 \\
2-3   & 50.000  & 16.667 & 16.667 & 16.667 & 0.000   & 0.000   & 0.000  & 0.219 & 0.193 & 20.056 & 0.078 \\
2-5   & 83.333  & 0.000  & 0.000  & 0.000  & 16.667  & 0.000   & 0.000  & 0.307 & 0.775 & 20.700 & 0.080 \\
2-7   & 100.000 & 0.000  & 0.000  & 0.000  & 0.000   & 0.000   & 0.000  & 0.264 & 0.524 & 20.411 & 0.079 \\
2-9   & 33.333  & 0.000  & 50.000 & 16.667 & 0.000   & 0.000   & 0.000  & 0.113 & 0.214 & 20.333 & 0.079 \\
2-11  & 83.333  & 0.000  & 0.000  & 0.000  & 16.667  & 0.000   & 0.000  & 0.183 & 0.785 & 21.533 & 0.080 \\
2-13  & 83.333  & 0.000  & 0.000  & 0.000  & 16.667  & 0.000   & 0.000  & 0.222 & 0.709 & 20.380 & 0.079 \\
2-15  & 50.000  & 0.000  & 0.000  & 0.000  & 33.333  & 0.000   & 16.667 & 0.140 & 0.534 & 20.767 & 0.079 \\
3-1   & 83.333  & 0.000  & 0.000  & 0.000  & 16.667  & 0.000   & 0.000  & 0.319 & 0.708 & 21.087 & 0.078 \\
3-3   & 66.667  & 0.000  & 16.667 & 0.000  & 16.667  & 0.000   & 0.000  & 0.175 & 0.217 & 20.758 & 0.079 \\
3-5   & 66.667  & 0.000  & 0.000  & 0.000  & 33.333  & 0.000   & 0.000  & 0.166 & 0.143 & 20.692 & 0.080 \\
3-7   & 66.667  & 0.000  & 16.667 & 0.000  & 16.667  & 0.000   & 0.000  & 0.233 & 0.546 & 21.492 & 0.081 \\
3-9   & 16.667  & 0.000  & 16.667 & 0.000  & 33.333  & 0.000   & 33.333 & 0.121 & 0.031 & 22.833 & 0.079 \\
3-11  & 33.333  & 0.000  & 50.000 & 0.000  & 0.000   & 0.000   & 16.667 & 0.320 & 1.603 & 21.483 & 0.078 \\
3-13  & 83.333  & 0.000  & 0.000  & 0.000  & 0.000   & 0.000   & 16.667 & 0.137 & 0.245 & 22.420 & 0.078 \\
3-15  & 66.667  & 0.000  & 0.000  & 0.000  & 0.000   & 16.667  & 16.667 & 0.120 & 0.148 & 23.717 & 0.078 \\
\textbf{Avg}   & \textbf{63.542} & \textbf{1.042} & \textbf{10.417} & \textbf{2.083} & \textbf{13.542} & \textbf{3.125} & \textbf{6.250} & \textbf{0.205} & \textbf{0.498} & \textbf{21.111} & \textbf{0.079} \\
\bottomrule
\end{tabular}
\end{table*}

\subsubsection{Details of Gen 2B testing}

The scene composition of Gen 2B has 16 more scenes compared to Gen 1C, ensuring that each of the additional scenes has sufficient parking route data. The model based on Gen 2B achieves the same TSR (77.083\%) as that of Gen 1C. Due to the shorter straight-line distance before braking in Gen 2, the model performs better on parking slots near the edge of the parking lot. In Gen 1C, the TSR for slots 2-1 and 3-1 is only 16.667\%, whereas in Gen 2B, the TSR for these slots is 66.667\% and 50\%, respectively.

The preliminary test result of model on Gen 2B is illustrated in Table~\ref{tab:Gen_2B_default}. The experiments consist of 96 tests in 16 different scenes. By observing the real-time video of the model's inference and control of the vehicle, it is found that the vehicle's speed is too high, leading to unintended movement towards non-target parking slots or even collisions. 

To address this issue, the speed for both gear 1 and reverse (R) was limited to 7.5 km/h. Additionally, a closed-loop experiment consisting of 384 test cases was conducted to evaluate the effectiveness of this adjustment. The result in Table~\ref{tab:Gen_2B_7.5} demonstrates that the model trained based on Gen 2B has a robust performance, achieving a TSR of 85.156\%. The model performs better in 7 tasks and maintains the same TSR in 3 tasks. Compared to the original speed limit set by the author, the stricter speed limit resulted in an increase of 1.417 seconds in APT. However, this adjustment led to a 6.69\% reduction in APE and a 14.76\% decrease in AOE. More importantly, CR decreased from 9.375\% to 2.865\%, improving safety.

\begin{table*}[h]
\small
\caption{Closed-loop results including 96 test cases for the model trained on Gen 2B, with the speed limit set as default.}\label{tab:Gen_2B_default}
\renewcommand\arraystretch{1.5}
\centering
\begin{tabular}{@{\hspace{0pt}}m{1.5cm}<{\centering}@{\hspace{0pt}}@{\hspace{0pt}}m{1.5cm}<{\centering}@{\hspace{0pt}}@{\hspace{0pt}}m{1.5cm}<{\centering}@{\hspace{0pt}}@{\hspace{0pt}}m{1.5cm}<{\centering}@{\hspace{0pt}}@{\hspace{0pt}}m{1.5cm}<{\centering}@{\hspace{0pt}}@{\hspace{0pt}}m{1.5cm}<{\centering}@{\hspace{0pt}}@{\hspace{0pt}}m{1.5cm}<{\centering}@{\hspace{0pt}}@{\hspace{0pt}}m{1.5cm}<{\centering}@{\hspace{0pt}}@{\hspace{0pt}}m{1.5cm}<{\centering}@{\hspace{0pt}}@{\hspace{0pt}}m{1.5cm}<{\centering}@{\hspace{0pt}}@{\hspace{0pt}}m{1.5cm}<{\centering}@{\hspace{0pt}}@{\hspace{0pt}}m{1.5cm}<{\centering}@{\hspace{0pt}}}
\toprule
TaskIdx & TSR (\%) & TFR (\%) & NTSR (\%) & NTFR (\%) & CR (\%) & OR (\%) & TR (\%) & APE (m) & AOE (deg) & APT (s) & AIT (s) \\
\midrule
2-1   & 66.667  & 0.000  & 0.000  & 0.000  & 0.000  & 33.333  & 0.000  & 0.277  & 0.821  & 20.458  & 0.078  \\
2-3   & 100.000 & 0.000  & 0.000  & 0.000  & 0.000  & 0.000   & 0.000  & 0.278  & 0.209  & 19.844  & 0.078  \\
2-5   & 83.333  & 0.000  & 0.000  & 0.000  & 16.667 & 0.000   & 0.000  & 0.273  & 0.315  & 20.673  & 0.078  \\
2-7   & 66.667  & 16.667 & 0.000  & 0.000  & 16.667 & 0.000   & 0.000  & 0.259  & 0.159  & 19.383  & 0.078  \\
2-9   & 50.000  & 0.000  & 0.000  & 0.000  & 50.000 & 0.000   & 0.000  & 0.242  & 0.313  & 18.711  & 0.078  \\
2-11  & 100.000 & 0.000  & 0.000  & 0.000  & 0.000  & 0.000   & 0.000  & 0.307  & 0.444  & 20.189  & 0.079  \\
2-13  & 100.000 & 0.000  & 0.000  & 0.000  & 0.000  & 0.000   & 0.000  & 0.299  & 0.611  & 19.417  & 0.078  \\
2-15  & 100.000 & 0.000  & 0.000  & 0.000  & 0.000  & 0.000   & 0.000  & 0.214  & 0.435  & 20.228  & 0.078  \\
3-1   & 50.000  & 0.000  & 0.000  & 0.000  & 0.000  & 33.333  & 16.667 & 0.213  & 1.218  & 23.489  & 0.078  \\
3-3   & 100.000 & 0.000  & 0.000  & 0.000  & 0.000  & 0.000   & 0.000  & 0.263  & 0.236  & 19.528  & 0.078  \\
3-5   & 100.000 & 0.000  & 0.000  & 0.000  & 0.000  & 0.000   & 0.000  & 0.286  & 0.180  & 20.711  & 0.079  \\
3-7   & 66.667  & 16.667 & 0.000  & 0.000  & 0.000  & 0.000   & 16.667 & 0.207  & 0.256  & 20.708  & 0.087  \\
3-9   & 33.333  & 0.000  & 0.000  & 0.000  & 66.667 & 0.000   & 0.000  & 0.209  & 0.382  & 19.950  & 0.090  \\
3-11  & 100.000 & 0.000  & 0.000  & 0.000  & 0.000  & 0.000   & 0.000  & 0.279  & 0.289  & 21.094  & 0.084  \\
3-13  & 66.667  & 0.000  & 0.000  & 0.000  & 0.000  & 0.000   & 33.333 & 0.242  & 0.239  & 22.900  & 0.079  \\
3-15  & 50.000  & 0.000  & 0.000  & 0.000  & 0.000  & 0.000   & 50.000 & 0.225  & 0.175  & 23.322  & 0.078  \\
\textbf{Avg}   & \textbf{77.083}  & \textbf{2.083}  & \textbf{0.000}  & \textbf{0.000}  & \textbf{9.375}  & \textbf{4.167}   & \textbf{7.292}  & \textbf{0.254}  & \textbf{0.393}  & \textbf{20.663}  & \textbf{0.080}  \\
\bottomrule
\end{tabular}
\end{table*}

\begin{table*}[htbp]
\small
\caption{Closed-loop results including 384 test cases for model trained on Gen 2B, with the speed limit set as 7.5 km/h.}\label{tab:Gen_2B_7.5}%
\renewcommand\arraystretch{1.5}
\centering
\begin{tabular}{@{\hspace{0pt}}m{1.5cm}<{\centering}@{\hspace{0pt}}@{\hspace{0pt}}m{1.5cm}<{\centering}@{\hspace{0pt}}@{\hspace{0pt}}m{1.5cm}<{\centering}@{\hspace{0pt}}@{\hspace{0pt}}m{1.5cm}<{\centering}@{\hspace{0pt}}@{\hspace{0pt}}m{1.5cm}<{\centering}@{\hspace{0pt}}@{\hspace{0pt}}m{1.5cm}<{\centering}@{\hspace{0pt}}@{\hspace{0pt}}m{1.5cm}<{\centering}@{\hspace{0pt}}@{\hspace{0pt}}m{1.5cm}<{\centering}@{\hspace{0pt}}@{\hspace{0pt}}m{1.5cm}<{\centering}@{\hspace{0pt}}@{\hspace{0pt}}m{1.5cm}<{\centering}@{\hspace{0pt}}@{\hspace{0pt}}m{1.5cm}<{\centering}@{\hspace{0pt}}@{\hspace{0pt}}m{1.5cm}<{\centering}@{\hspace{0pt}}}
\toprule
TaskIdx & TSR (\%) & TFR (\%) & NTSR (\%) & NTFR (\%) & CR (\%) & OR (\%) & TR (\%) & APE (m) & AOE (deg) & APT (s) & AIT (s) \\
\midrule
2-1   & 54.167  & 0.000  & 8.333  & 0.000  & 0.000  & 20.833 & 16.667 & 0.259 & 0.586 & 22.645 & 0.079 \\
2-3   & 91.667  & 0.000  & 0.000  & 0.000  & 0.000  & 0.000  & 8.333  & 0.258 & 0.193 & 22.283 & 0.079 \\
2-5   & 100.000 & 0.000  & 0.000  & 0.000  & 0.000  & 0.000  & 0.000  & 0.253 & 0.182 & 21.131 & 0.079 \\
2-7   & 100.000 & 0.000  & 0.000  & 0.000  & 0.000  & 0.000  & 0.000  & 0.257 & 0.141 & 21.233 & 0.079 \\
2-9   & 70.833  & 0.000  & 4.167  & 0.000  & 20.833 & 0.000  & 4.167  & 0.250 & 0.207 & 21.748 & 0.079 \\
2-11  & 100.000 & 0.000  & 0.000  & 0.000  & 0.000  & 0.000  & 0.000  & 0.239 & 0.450 & 21.347 & 0.081 \\
2-13  & 100.000 & 0.000  & 0.000  & 0.000  & 0.000  & 0.000  & 0.000  & 0.309 & 0.360 & 20.854 & 0.080 \\
2-15  & 95.833  & 0.000  & 0.000  & 0.000  & 0.000  & 0.000  & 4.167  & 0.262 & 0.481 & 22.539 & 0.079 \\
3-1   & 62.500  & 0.000  & 0.000  & 0.000  & 0.000  & 37.500 & 0.000  & 0.245 & 0.868 & 21.551 & 0.079 \\
3-3   & 95.833  & 0.000  & 0.000  & 0.000  & 4.167  & 0.000  & 0.000  & 0.168 & 0.324 & 21.498 & 0.079 \\
3-5   & 91.667  & 0.000  & 0.000  & 0.000  & 0.000  & 0.000  & 8.333  & 0.306 & 0.505 & 22.469 & 0.080 \\
3-7   & 95.833  & 0.000  & 0.000  & 0.000  & 0.000  & 0.000  & 4.167  & 0.190 & 0.134 & 21.637 & 0.080 \\
3-9   & 75.000  & 0.000  & 0.000  & 0.000  & 20.833 & 0.000  & 4.167  & 0.237 & 0.232 & 21.693 & 0.080 \\
3-11  & 100.000 & 0.000  & 0.000  & 0.000  & 0.000  & 0.000  & 0.000  & 0.196 & 0.171 & 22.703 & 0.079 \\
3-13  & 62.500  & 0.000  & 0.000  & 0.000  & 0.000  & 0.000  & 37.500 & 0.122 & 0.255 & 24.681 & 0.079 \\
3-15  & 66.667  & 0.000  & 0.000  & 0.000  & 0.000  & 0.000  & 33.333 & 0.242 & 0.266 & 23.268 & 0.079 \\
\textbf{Avg} & \textbf{85.156} & \textbf{0.000} & \textbf{0.781} & \textbf{0.000} & \textbf{2.865} & \textbf{3.646} & \textbf{7.552} & \textbf{0.237} & \textbf{0.335} & \textbf{22.080} & \textbf{0.079} \\
\bottomrule
\end{tabular}
\end{table*}

\subsubsection{Parking performance compared with baseline and original expert}

According to Table~\ref{comparison_baseline_expert}, the model trained on the new dataset Gen 2B can essentially reproduce the performance of the baseline. Although TSR is 6.25 percentage points lower and APT increases by 6.36 seconds, APE decreases by 20\%, and AOE decreases by 60.9\%. Meanwhile, the new expert shows a significant improvement in accuracy compared to the original expert. Although APT increases by 1.99 seconds, APE decreases by 43.5\% and AOE decreases by 90.0\%.

\begin{table*}[htbp]
\small
\caption{Performance comparison with baseline and original expert}\label{comparison_baseline_expert}%
\renewcommand\arraystretch{1.5}
\centering
\begin{tabular}{@{\hspace{0pt}}m{3cm}<{\centering}@{\hspace{0pt}}@{\hspace{0pt}}m{2.5cm}<{\centering}@{\hspace{0pt}}@{\hspace{0pt}}m{2.5cm}<{\centering}@{\hspace{0pt}}@{\hspace{0pt}}m{2.5cm}<{\centering}@{\hspace{0pt}}@{\hspace{0pt}}m{2.5cm}<{\centering}@{\hspace{0pt}}@{\hspace{0pt}}m{2.5cm}<{\centering}@{\hspace{0pt}}@{\hspace{0pt}}m{2.5cm}<{\centering}@{\hspace{0pt}}}
\toprule
TaskIdx & TSR (\%)  & TFR (\%)  & CR (\%)  & APE (m)  & AOE (deg) & APT (s) \\
\midrule
Baseline & 91.41 & 2.08 & 2.08 & 0.30 & 0.87 & 15.72 \\
Model on Gen 2B   & 85.16 & 0.00 & 2.87 & 0.24 & 0.34 & 22.08 \\
Original Expert   & 100.00 & 0.00 & 0.00 & 0.23 & 0.48 & 14.96 \\
New Expert &  100.00 & 0.00 & 0.00 & 0.13 & 0.05 & 16.95 \\
\bottomrule
\end{tabular}
\end{table*}

\begin{table*}[htbp]
\caption{Comparison of GEN 1C's Performance under Different Initial Learning Rates}
\renewcommand\arraystretch{1.5}
\label{tab:Gen_1C_comparison}
\centering
\begin{tabular}{@{\hspace{0pt}}m{3cm}<{\centering}@{\hspace{0pt}}@{\hspace{0pt}}m{2.5cm}<{\centering}@{\hspace{0pt}}@{\hspace{0pt}}m{2.5cm}<{\centering}@{\hspace{0pt}}@{\hspace{0pt}}m{2.5cm}<{\centering}@{\hspace{0pt}}@{\hspace{0pt}}m{2.5cm}<{\centering}@{\hspace{0pt}}@{\hspace{0pt}}m{2.5cm}<{\centering}@{\hspace{0pt}}@{\hspace{0pt}}m{2.5cm}<{\centering}@{\hspace{0pt}}}
\toprule
\multirow{2}{*}{\textbf{Metric}} & \multicolumn{3}{c}{\textbf{Lr = $1 \times 10^{-4}$}} & \multicolumn{3}{c}{\textbf{Lr = $7.5 \times 10^{-5}$}} \\
\cmidrule(lr){2-4} \cmidrule(lr){5-7}
 & Epochs & val\_loss & TSR(\%) & Epochs & val\_loss & TSR(\%) \\
\midrule
 \textbf{Early} & -- & -- & -- & 65 & 0.799 & 73.958 \\
 \textbf{Mid} & 150 & 0.752 & 72.917 & 155 & 0.764 & 77.083 \\
 \textbf{Late} & 190 & 0.746 & 72.917 & 200 & 0.757 & 75.000 \\
\bottomrule
\end{tabular}
\end{table*}

\begin{table*}[htbp]
\caption{Comparison of Gen 2A and Gen 2B's Performance}\label{tab:Gen_2A_2B}%
\renewcommand\arraystretch{1.5}
\centering
\begin{tabular}{@{\hspace{0pt}}m{3cm}<{\centering}@{\hspace{0pt}}@{\hspace{0pt}}m{2.8cm}<{\centering}@{\hspace{0pt}}@{\hspace{0pt}}m{2.8cm}<{\centering}@{\hspace{0pt}}}
\toprule
\textbf{Metric} & \textbf{Gen 2A} & \textbf{Gen 2B}  \\
\midrule
Epochs         & 150    & 130 \\
val\_loss & 0.9550 & 0.8089 \\
TSR(\%)       & 63.542 & 77.083 \\
\bottomrule
\end{tabular}
\footnotetext{The initial learning rate is set as $7.5 \times 10^{-5}$ during training on both datasets.}
\end{table*}

\section{Conclusion}\label{Conclusion}

In this work we created an open-source dataset for E2E Parking and achieved a Target Success Rate outperforming the prior work. Moreover, the model trained on the dataset shows an enhanced parking precision with lower Average
Position Error and Average Orientation Error. The iterative process of dataset construction provides insights for relevant researchers on how to optimize the dataset and improve training and testing results through methods such as parking operation strategies, scene selection, learning rate adjustment, and speed limitations.

Despite the promising results, this work has several limitations. First, the dataset was collected exclusively under optimal lighting conditions (sunny days at noon), which limits its ability to generalize to scenarios with different times of day or adverse weather conditions. Second, all obstacles present in the dataset are static, and therefore the model has not been exposed to dynamic objects, which may hinder its capacity for motion prediction and collision avoidance in real-world environments. Third, the current experiments are conducted in a simulated environment, which inevitably introduces a domain gap; the proposed algorithm may not transfer seamlessly to real-world applications due to sensor noise, actuation delays, and other environmental uncertainties.

Nonetheless, we believe this work provides a solid foundation for future research in end-to-end autonomous parking. By expanding the dataset to include more diverse environmental conditions and dynamic agents, refining the model architecture, and adapting and validating the proposed algorithm on real vehicles, future work can further advance the deployment of robust autonomous parking systems.


\bibliographystyle{IEEEtran}
\bibliography{sn-bibliography}


\end{document}